\documentclass{article}

\usepackage{nac_preprint}

\usepackage[utf8]{inputenc} 
\usepackage[T1]{fontenc}    
\usepackage{hyperref}       
\usepackage{url}            
\usepackage{booktabs}       
\usepackage{amsfonts}       
\usepackage{nicefrac}       
\usepackage{microtype}      
\usepackage{xcolor}         
\usepackage{tcolorbox}

\usepackage{booktabs}
\usepackage{subcaption}
\usepackage{amsmath}
\usepackage{mathtools}
\usepackage[toc,page]{appendix}
\usepackage{enumitem}
\usepackage{graphics}
\usepackage{graphicx}
\usepackage[export]{adjustbox}
\usepackage{algorithm}
\usepackage{algorithmicx}
\usepackage[noend]{algpseudocode}
\algnewcommand{\LineComment}[1]{\State \(//\) #1}

\usepackage{wrapfig}

\usepackage{graphicx}
\usepackage{url}
\usepackage{subcaption}
\usepackage{amsmath,amsfonts,bbm,dsfont}
\usepackage{placeins}

\title{Maze Learning using a Hyperdimensional Predictive Processing Cognitive Architecture}
\author{%
  Alexander Ororbia$^\dagger$ \\
  \texttt{ago@cs.rit.edu} \\
  \And
  M. Alex Kelly$^\ddagger$ \\
  \texttt{alex.kelly@carleton.ca} 
  \AND
  \vspace{-0.75cm}\\
  $^\dagger$ Rochester Institute of Technology, Rochester, NY, 14623, USA. \\
  $^\ddagger$ Carleton University, Ottawa, ON, K1S 5B6, Canada. 
}

\begin{document}

\maketitle

\begin{abstract}
We present the COGnitive Neural GENerative system (CogNGen), a cognitive architecture that combines two neurobiologically-plausible, computational models: predictive processing and hyperdimensional/vector-symbolic models. We draw inspiration from architectures such as ACT-R and Spaun/Nengo. CogNGen is in broad agreement with these, providing a level of detail between ACT-R’s high-level symbolic description of human cognition and Spaun’s low-level neurobiological description, furthermore creating the groundwork for designing agents that learn continually from diverse tasks and model human performance at larger scales than what is possible with current systems. We test CogNGen on four maze-learning tasks, including those that test memory and planning, and find that CogNGen matches performance of deep reinforcement learning models and exceeds on a task designed to test memory. 

\end{abstract}

\section{Introduction}
\label{sec:intro}

Artificial neural networks (ANNs) do not typically model high-level cognition and are usually models of only one task. Otherwise, when an ANN is trained to learn a series of tasks, catastrophic interference occurs, with each new task causing the ANN to forget all prior tasks \cite{french1999catastrophic,mannering2021catastrophic,mccloskey_catastrophic_1989}. On the other hand, symbolic cognitive architectures, such as the widely used ACT-R \cite{Ritter2019actr}, can capture the complexities of high-level cognition but scale poorly to the naturalistic data of sensory perception or to big data necessary for modelling life-long learning. 

We propose a cognitive architecture \cite{Ororbia2022cogsci} that is built from two neurobiologically and cognitively plausible models, namely neural generative coding (NGC) \cite{ororbia2022neural} (a form of predictive processing) 
and vector-symbolic (a.k.a. hyperdimensional) models of memory \cite{Hintzman1984,KellyMewhortWest}.
Desirably, using these specific building blocks yields scalable, local Hebbian \cite{hebb1949organization} update rules for adjusting the system's synapses while facilitating robustness in acquiring, storing, and composing representations of tasks encountered sequentially \cite{mannering2021catastrophic}.
Our intent is to advance towards an architecture capable of intelligent action at all scales of learning, from the small maze tasks considered here, to skills acquired gradually over a lifetime. By combining NGC with vector-symbolic models of human memory, we work towards creating a model of cognition that has the power of modern machine learning techniques while retaining long-term memory, single-trial and transfer-learning, planning, and other capacities associated with high-level cognition.

In this work, we demonstrate proof of concept and show that our architecture, CogNGen (the COGnitive Neural GENerative system; see \cite{Ororbia2022cogsci} for details), can learn variants of a maze-learning task, including those requiring planning (get a key to open a locked door) and memory (pick a path based on an earlier cue). Our results show that CogNGen is competitive with several deep learning approaches, offering promising performance when task reward is sparse. We start by describing the circuits and core modules used to construct CogNGen. Then, we describe the tasks used to evaluate CogNGen and the experimental results. 

\section{Neural Building Blocks}
\label{sec:building_blocks}




\subsection{Neural Generative Coding (NGC)}
\label{sec:ngc}

Neural generative coding (NGC) is an instantiation of the predictive processing brain theory \cite{rao1999predictive,friston2005theory}, yielding a robust form of predict-then-correct learning and inference. 
An NGC circuit in CogNGen receives two sensory vectors, input $\mathbf{x}^i \in \mathcal{R}^{I \times 1}$ ($I$ is the input dimensionality) and  output $\mathbf{x}^o \in \mathcal{R}^{O \times 1}$ ($O$ is the output dimensionality). An NGC circuit is composed of $L$ layers of neurons, i.e., layer $\ell$ is represented by state vector $\mathbf{z}^\ell \in \mathcal{R}^{H_\ell \times 1}$ containing $H_\ell$ total units. Given an input--output pair of sensory vectors $\mathbf{x}^i$ and $\mathbf{x}^o$, the circuit clamps the last layer $\mathbf{z}^L$ to the input, $\mathbf{z}^L =\mathbf{x}^i$, and clamps the first layer $\mathbf{z}^0$ to the output, $\mathbf{z}^0 =\mathbf{x}^o$. Once clamped, the NGC circuit will undergo a settling cycle where it processes the input and output vectors for several steps in time (i.e., it processes sensory signals over a stimulus window of $K$ discrete time steps). 
After processing the input--output pair over a stimulus window, the synaptic matrices are adjusted via local Hebbian-like updates. See the Appendix\footnote{Appendix: \url{https://www.cs.rit.edu/$\sim$ago/cogngnen\_agi2022\_append.pdf}} for details of the exact mechanics/dynamics of the NGC circuits we implemented for this paper.

\subsection{Memory}

For CogNGen, we model both short and long-term memory using the MINERVA~2 model of human memory \cite{Hintzman1984}. Short-term MINERVA~2 is cleared after an episode is completed (e.g., a maze is solved), whereas the contents of long-term MINERVA~2 persist across episodes. MINERVA~2 is a model of human memory equivalent to a type of Hebbian network \cite{KellyMewhortWest}. We choose MINERVA~2 since it captures a wide variety of human memory phenomena, e.g., \cite{Hintzman1984,Kelly2020indirect,KellyMewhortWest}. Our implementation of MINERVA~2 stores a sequence of observations as a concatenated vector. Each sequence is represented as a row in the memory table. Retrieval from memory is a weighted sum of all rows in the table, each row weighted by the similarity to the currently observed sequence, allowing MINERVA~2 to predict the next observations(s) given the agent's recent history. Growth of the memory table is limited by forgetting simulated as random deletion \cite{Hintzman1984}. 

\section{The CogNGen Cognitive Architecture}
\label{sec:arch}


\subsection{Perceptual Modules}

CogNGen's perceptual module encodes observation $\mathbf{o}_t \in \mathcal{R}^{D_o \times 1}$ at time $t$ to $\mathbf{z}_t \in \mathcal{R}^{D_z \times 1}$ (and decodes it back) -- $D_o$ is the dimension of $\mathbf{o}_t$ and $D_z$ is that of $\mathbf{z}_t$.  
Although this process can be implemented in NGC circuits, in this work, we leverage an encoder and decoder offered by the task environment (see Appendix).

\subsection{Procedural Memory and Motor Control}
\label{sec:procedural}

\paragraph{The Procedural Dynamics Model: }
Motivated by the finding of expected value estimation in the brain, CogNGen's procedural module implements a neural circuit that produces intrinsic reward signals. At a high level, this neural machinery facilitates some of the functionality of the basal ganglia and procedural memory, simulating an internal reward-creation process \cite{schultz2016reward}.
Concretely, we refer to the above as an NGC dynamics model, where reward is calculated as a function of its error neurons, further coupled to a short-term MINERVA~2 memory ``filter''. 

The NGC dynamics circuit processes the current state $\mathbf{z}_t$ and the external discrete action $\mathbf{a}^{ext}_t$ ($\mathbf{a}^{ext}_t \in \{0,1\}^{A_{ext} \times 1}$ is its one-hot encoding, where $A_{ext}$ is the number of actions), as produced by the motor-action model (described later), and predicts the value of the future state $\mathbf{z}_{t+1}$. When provided with $\mathbf{z}_{t+1}$, the dynamics circuit runs the following for its layer-wise predictions:
\begin{align}
    \mathbf{\bar{z}}^2 & = \mathbf{W}^3_{ext} \cdot \mathbf{a}^{ext}_t + \mathbf{W}^3_z \cdot \mathbf{z}_t + \mathbf{b}_2 \\ 
    \mathbf{\bar{z}}^1 & = \mathbf{W}^2 \cdot \phi( \mathbf{z}^2_t ) + \mathbf{b}_1 \\
    \mathbf{\hat{z}}_{t+1} = \mathbf{\bar{z}}^0 & = g^0\Big(\mathbf{W}^1 \cdot \phi( \mathbf{z}^1_t ) + \mathbf{b}_0 \Big)
\end{align}
and leverages the NGC settling process (see Appendix) to compute its internal state values, i.e., $\mathbf{z}^3_t, \mathbf{z}^2_t, \mathbf{z}^1_t$. 
Notice that we have simplified a few items with respect to the NGC circuit -- the topmost layer-wise prediction $\mathbf{\bar{z}}^3_t$ sets $\phi^3(\mathbf{v}) = \mathbf{v}$ for both its top-most inputs $\mathbf{c}^{ext}_t$ and $\mathbf{z}_t$, the post-activation prediction functions for the internal layers are $g^2(\mathbf{v}) = g^1(\mathbf{v}) = \mathbf{v}$, and  $phi^2(\mathbf{v}) = \phi^1(\mathbf{v}) = \phi(\mathbf{v})$ (the same state activation function type is used in calculating $\mathbf{\hat{z}}^1$ and $\mathbf{\hat{z}}^0$). Once the above dynamics have been executed, the NGC dynamics model's synapses are adjusted via Hebbian updates. 
Furthermore, upon receiving $\mathbf{z}_{t+1}$, the short-term MINERVA~2 coupled to the dynamics circuit stores the current latent state vector, updating its current knowledge about the episode that CogNGen is operating with, and outputs a similarity score $s^{recall}$. Note that, at the an episode's termination, the contents of the short-term MINERVA~2 are cleared.

To generate the value of the epistemic reward \cite{ororbia2022backpropfree}), the dynamics model first settles to a prediction $\mathbf{\hat{z}}_{t+1}$ given the value of CogNGen's next latent state $\mathbf{z}_{t+1}$. After its settling process has finished, the activity signals of its (squared) error neurons are summed to obtain the circuit's  epistemic reward signal:
\begin{align}
    r^{ep}_t &= \sum_j (\mathbf{e}^0 )^2_{j,1} + \sum_j (\mathbf{e}^1 )^2_{j,1} + \sum_j (\mathbf{e}^2 )^2_{j,1} \\
    r^{ep}_t &= r^{ep}_t / (r^{ep}_{max}) \quad \mbox{where } r^{ep}_{max} = \max(r^{ep}_1, r^{ep}_2, ..., r^{ep}_t)
\end{align}
where the epistemic reward signal is normalized to the range of $[0,1]$ by tracking the maximum epistemic signal observed throughout the course of the simulation. This signal is next modified by the MINERVA~2 memory filter as follows: 
\begin{align}
    r^{ep} =
  \begin{cases} 
      \eta_e r^{ep} & s^{recall} \leq s_\theta \\
      -0.1 & \mbox{otherwise}
  \end{cases}
\end{align}
where $s_\theta$ is a threshold that $s^{recall}$ is compared against and $0 \leq \eta_e \leq 1$ is meant to weight the epistemic signal. If $s^{recall} \leq s^\theta$, then $\mathbf{z}_{t+1}$ is deemed ``unfamiliar'' and the agent is positively rewarded with the epistemic reward for uncovering a new state of its environment. Whereas if the opposite is true ($s^{recall} > s^{\theta}$), then the latent state is deemed familiar and the agent is given a negative penalty. 
The final reward signal is computed by combining the epistemic signal with the problem-specific (instrumental) reward: $r^{in}_t$, i.e., $r_t = r^{in}_t + r^{ep}_t$.  Although we utilize the sparse reward signal provided by the task for $r^{in}_t$, we remark that another circuit, serving as CogNGen's prior preference could be designed to encode 
probability distributions over preferred goal states \cite{friston2017active,ororbia2022backpropfree}. 

\paragraph{The Motor Action Model: }

To manipulate its environment, CogNGen implements another NGC circuit that we call the motor-action model $f_a \colon \mathbf{z}_t \mapsto (\mathbf{c}^{int}_t, \mathbf{c}^{ext}_t)$ (offering some functionality provided by the motor cortex) which outputs two control signals at each time step, i.e., internal control signal $\mathbf{c}^{int}_t \in \mathcal{R}^{A_int \times 1}$ and external control signal $\mathbf{c}^{ext}_t \in \mathcal{R}^{A_ext \times 1}$. 
Note that a discrete internal action $a^{int}_t \in \{1,2,,...,i,...,A_{int}\}$ is extracted via $a^{int}_t = \arg\max_{i} \mathbf{c}^{int}_t$ and external action $a^{ext}_t \in \{1,2,,...,j,...,A_{ext}\}$ is extracted via $a^{ext}_t = \arg\max_{j} \mathbf{c}^{ext}_t$  ($A_{int}$ is the number of discrete internal actions).
Action $a^{ext}_t$ affects the environment while 
action $a^{int}_t$ manipulates the action model's coupled working memory buffers.

Within the NGC action-motor model is a modifiable working memory that allows the model to store a finite quantity $M_w$ of latent state vectors into a set of self-recurrent memory vector slots. This particular working memory module, which we call the \textit{self-recurrent slot buffer} 
serves as the glue that joins the modules of CogNGen together. 
The buffers in CogNGen serve the same purpose as ACT-R's buffers \cite{Ritter2019actr}. 
Each memory slot in the buffer is represented by $\mathbf{m}^i \in \mathcal{R}^{M_d \times 1}$ ($M_d$ is the dimesionality of the memory slot). This component of the action-motor model is inspired by the working memory model proposed in \cite{kruijne2021flexible}. Concretely, the self-recurrent slot buffer operates according to the following:
\begin{align}
    \mathbf{k}^i_t &= \mathbf{Q}^i \cdot \mathbf{z}_t, \forall i = 1,...,M_w \quad &\mbox{// Compute key} \label{eqn:key} \\
    s^i = \mathbf{s}^i &= \frac{1}{|\mathbf{m}^i|} \bigg( \sum_j \lfloor \mathbf{m}^i - \mathbf{k}^i_t \rfloor_{j,1} + \lfloor \mathbf{k}^i_t - \mathbf{m}^i \rfloor_{j,1} \bigg)  \quad &\mbox{// Compute match}  \label{eqn:match} \\
    \mathbf{m}_t &= \Big[ [\mathbf{m}^1,\mathbf{s}^1],...,[\mathbf{m}^i,\mathbf{s}^i],...,[\mathbf{m}^{M_w},\mathbf{s}^{M_w}] \Big] \quad &\mbox{// Compute value}  \label{eqn:value}
\end{align}
where $\mathbf{Q}^i \in \mathcal{R}^{M_d \times D_z}$ is the $i$th random projection matrix (sampled from a centered Gaussian distribution in this paper), which means there is one projection matrix per working memory slot. Note that the match score for any slot $i$ is $\mathbf{s}^i = \mathcal{R}^{1\times1}$ (a $1\times1$ vector) and thus also a scalar $s^i$. The working memory buffers, in essence, compute a key value vector $\mathbf{k}^i_t$ given the current state input $\mathbf{z}_t$ for each slot (by projecting via matrix $\mathbf{Q}^i$), calculate the match score between the $i$th key and $i$th slot/value, and then return the entire concatenated contents $\mathbf{m}_t$ of working memory (including the match scores).

Given the output of working memory $\mathbf{m}_t$, the motor-action model then proceeds to compute its output control signals using an ancestral projection scheme (see Appendix), yielding $\mathbf{c}^{ext}_t, \mathbf{c}^{int}_t = f_{proj}(\mathbf{z}_t; \Theta)$, implemented as follows: 
\begin{align}
    \mathbf{\bar{z}}^3_t &= \mathbf{W}^4 \cdot \mathbf{z}_t + \phi( \mathbf{M} \cdot \mathbf{m}_t ) + \mathbf{b}^3 \label{eqn:motor_lyr3} \\
    \mathbf{\bar{z}}^2_t &= \mathbf{W}^3 \cdot \phi( \mathbf{z}^3_t ) + \mathbf{b}^2  \label{eqn:motor_lyr2} \\
    \mathbf{\bar{z}}^1_t &= \mathbf{W}^2 \cdot \phi( \mathbf{z}^2_t ) + \mathbf{b}^1  \label{eqn:motor_lyr1} \\
    \mathbf{c}^{ext}_t = \mathbf{\bar{z}}^0_{t,ext} &=  \mathbf{W}^1_{ext} \cdot  \phi( \mathbf{z}^1_t ) + \mathbf{b}^0_{ext}  \label{eqn:motor_lyr0_ext} \\
    \mathbf{c}^{int}_t = \mathbf{\bar{z}}^0_{t,int} &=  \mathbf{W}^1_{int} \cdot  \phi( \mathbf{z}^1_t ) + \mathbf{b}^0_{int} \mbox{.}  \label{eqn:motor_lyr0_int} 
\end{align}

\noindent The NGC circuit depicted in Equations \ref{eqn:motor_lyr3}-\ref{eqn:motor_lyr0_int} embodies both the ``internal control'' and ``control'' sub-systems by outputting $\mathbf{\bar{z}}^0_{t,ext}$, i.e., the same as control signal $\mathbf{c}^{ext}_t$, and $\mathbf{\bar{z}}^0_{t,int}$, i.e., the same as control signal $\mathbf{c}^{int}_t$.
The above dynamics represent a five-layer circuit with its top-most layer clamped to: $\mathbf{z}^4_t = \mathbf{z}_t$ and $\mathbf{m}_t$. 

Finally, after the motor-action model has produced its control signals, the internal action is selected via $a^{int}_t = \arg\max_{j} \mathbf{c}^{int}_t$ and the external action is selected via $a^{ext}_t = \arg\max_{j} \mathbf{c}^{ext}_t$. While $a^{ext}_t$ is transmitted to the environment, 
$a^{int}_t$ is used to modify the working memory module. The internal actions possible are specifically: $a^{int}_t = \{\mbox{ignore}, \mbox{store}_1, \mbox{store}_2,...,\mbox{store}_{M_w}\}$ (each integer has been mapped to a string clarifying the action's effect), where ``$\mbox{ignore}$'' means $\mathbf{z}_t$ is not stored and ``$\mbox{store}_i$'' means store $\mathbf{z}_t$ into memory slot $i$.

To update the motor-action model's synaptic efficacies, we then leverage the reward $r_t$ computed by the dynamics model described in Section \ref{sec:procedural}. Specifically, we compute the target control vectors $\mathbf{z}^0_{t,ext}$ and $\mathbf{z}^0_{t,int}$ as follows:
\begin{align}
    \mathbf{c}^{ext}_t, \mathbf{c}^{int}_t &= f_{proj}(\mathbf{z}_{t+1}; \Theta) \\
    z^0_{ext} &= \begin{cases} 
                    r_t & \mbox{if } \mathbf{z}_t \mbox{ is terminal} \\
                    r_t + \gamma \max_a \mathbf{c}^{ext}_t &   \mbox{otherwise }
                  \end{cases} \\
    z^0_{int} &= \begin{cases} 
                    r_t & \mbox{if } \mathbf{z}_t \mbox{ is terminal} \\
                    r_t + \gamma \max_a \mathbf{c}^{int}_t &   \mbox{otherwise }
                  \end{cases}            
\end{align}
and the final target vectors computed simply as:
\begin{align*}
    \mathbf{z}^0_{t,ext} &= z^0_{ext} \mathbf{a}^{ext}_t + (1 - \mathbf{a}^{ext}_t) \odot \mathbf{c}^{ext}_t \\
    \mathbf{z}^0_{t,int} &= z^0_{int} \mathbf{a}^{int}_t + (1 - \mathbf{a}^{int}_t) \odot \mathbf{c}^{int}_t \mbox{.}
\end{align*}
Once the target vectors have been created, the NGC settling process can be executed and all motor-action synapses are updated via Hebbian learning.


\subsection{Long-Term Memory}

CogNGen implements long-term memory through a MINERVA~2 module. Information is transferred to this memory through an intermediate working memory buffer, where pieces of a transition (partial experience) are stored as they are encountered during the agent-environment interaction process. Specifically, once the buffer contains at least one partial transition $(\mathbf{z}_t, \mathbf{a}^{ext}_t, \mathbf{a}^{int}_t, \mathbf{r}_t)$ ($\mathbf{r}_t \in \mathcal{R}^{1\times1}$)
, our long-term MINERVA~2 $\mathcal{M}$ (which is created alongside a starting transition buffer $\mathcal{S}_0$) 
is updated according to the following algorithm:
\begin{tcolorbox}[width=\linewidth, sharp corners=all,left=0pt,right=0pt,boxrule=0pt, colback=white!95!black]
\begin{enumerate}[noitemsep,nolistsep]
    \item Create a window (buffer) $w$ of length $L$ -- each slot is filled with empty values (zero vectors of the correct length). Store the start transition $\mathbf{m}^{exp}_0 = [\mathbf{z}_0, \mathbf{a}^{ext}_0, \mathbf{a}^{int}_0, \mathbf{r}_0]$ in buffer $\mathcal{S}_0$.
    \item Store $\mathbf{m}^{exp}_t = [\mathbf{z}_t, \mathbf{a}^{ext}_t, \mathbf{a}^{int}_t, \mathbf{r}_t]$ at the last position (index $L$) of the window $w$ and delete the entry at position $0$.
    \item Flatten $w$ into a vector $\mathbf{w}_{mem}$ and store this item by updating $\mathcal{M}$.
    \item If episode terminal has been reached, go to Step 1, else go to Step 2.
\end{enumerate}
\end{tcolorbox}
The above process is repeated until the end of simulation. We impose an upper bound on the number of transitions stored in $\mathcal{M}$ 
-- if this bound is exceeded, we remove the earliest transition $\mathbf{m}^{exp}_t$ stored in $\mathcal{M}$ and update $\mathcal{S}_0$ accordingly.

To drive learning through experience replay, CogNGen samples from $\mathcal{M}$ by:
\begin{tcolorbox}[width=\linewidth, sharp corners=all,left=0pt,right=0pt,boxrule=0pt, colback=white!95!black]
\begin{enumerate}[noitemsep,nolistsep]
    \item Create window $w$ of length $L$, initialized with empty values. Sample $\mathbf{m}^{exp}_0 \sim \mathcal{S_0}$ and place it in the last position $L$ in $w$.
    \item Remove the item at position $1$ in $w$ and use $\mathcal{M}$ to hetero-associatively complete/predict $\mathbf{m}^{exp}_{t+1}$.
    \item Store $\mathbf{m}^{exp}_{t+1}$ at last position $L$ within $w$.
    \item Repeat steps 2 through 4 until episode terminal reached. 
\end{enumerate}
\end{tcolorbox}
The above is repeated until $E$ episodes have been sampled. To create a mini-batch for updating the motor-action/dynamics circuits, we sample $B$ transitions $(\mathbf{z}_j, \mathbf{a}^{ext}_j, \mathbf{a}^{int}_j, \mathbf{r}_t,  \mathbf{z}_{j+1})$ from each sampled episode. Thus, at $t$, CogNGen's computation consists of an information processing step followed by a learning step.

\section{Experimental Results}
\label{sec:experiments}

\begin{table}[t!]
\begin{center}
\begin{tabular}{c c c c}
    \includegraphics[height=0.850in]{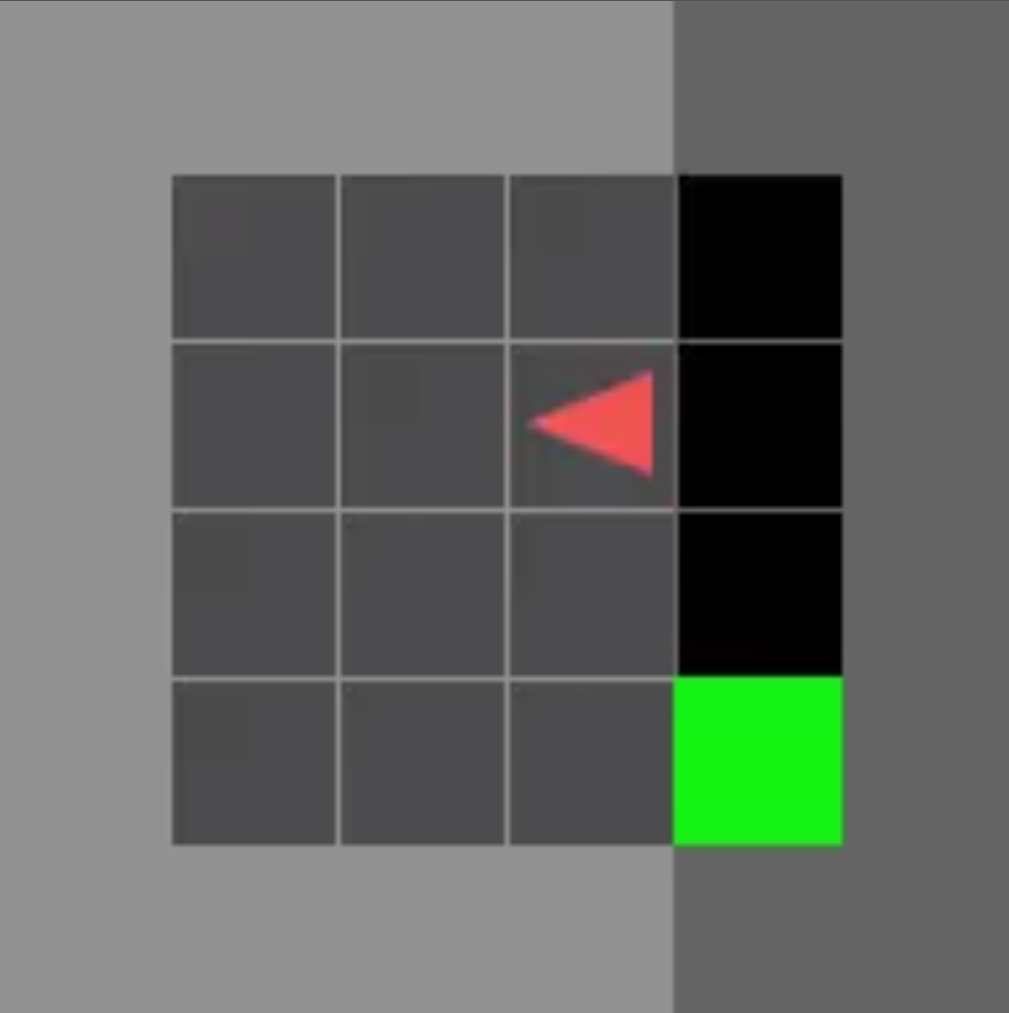} & \includegraphics[height=0.850in]{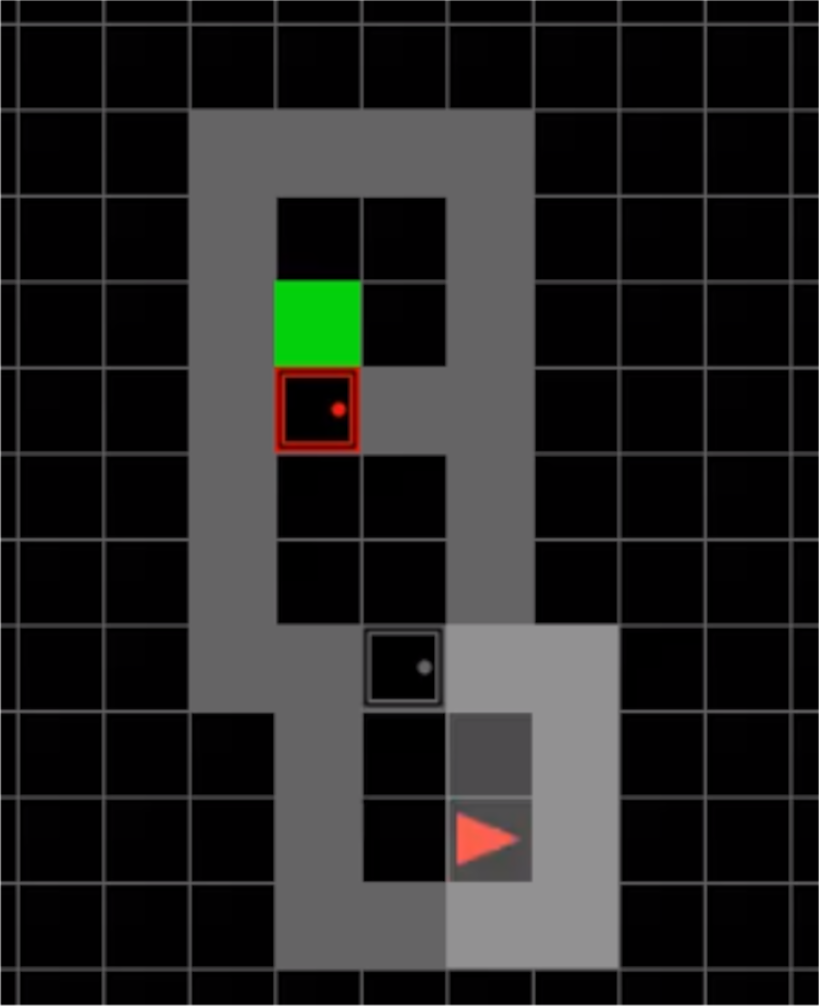} & \includegraphics[height=0.850in]{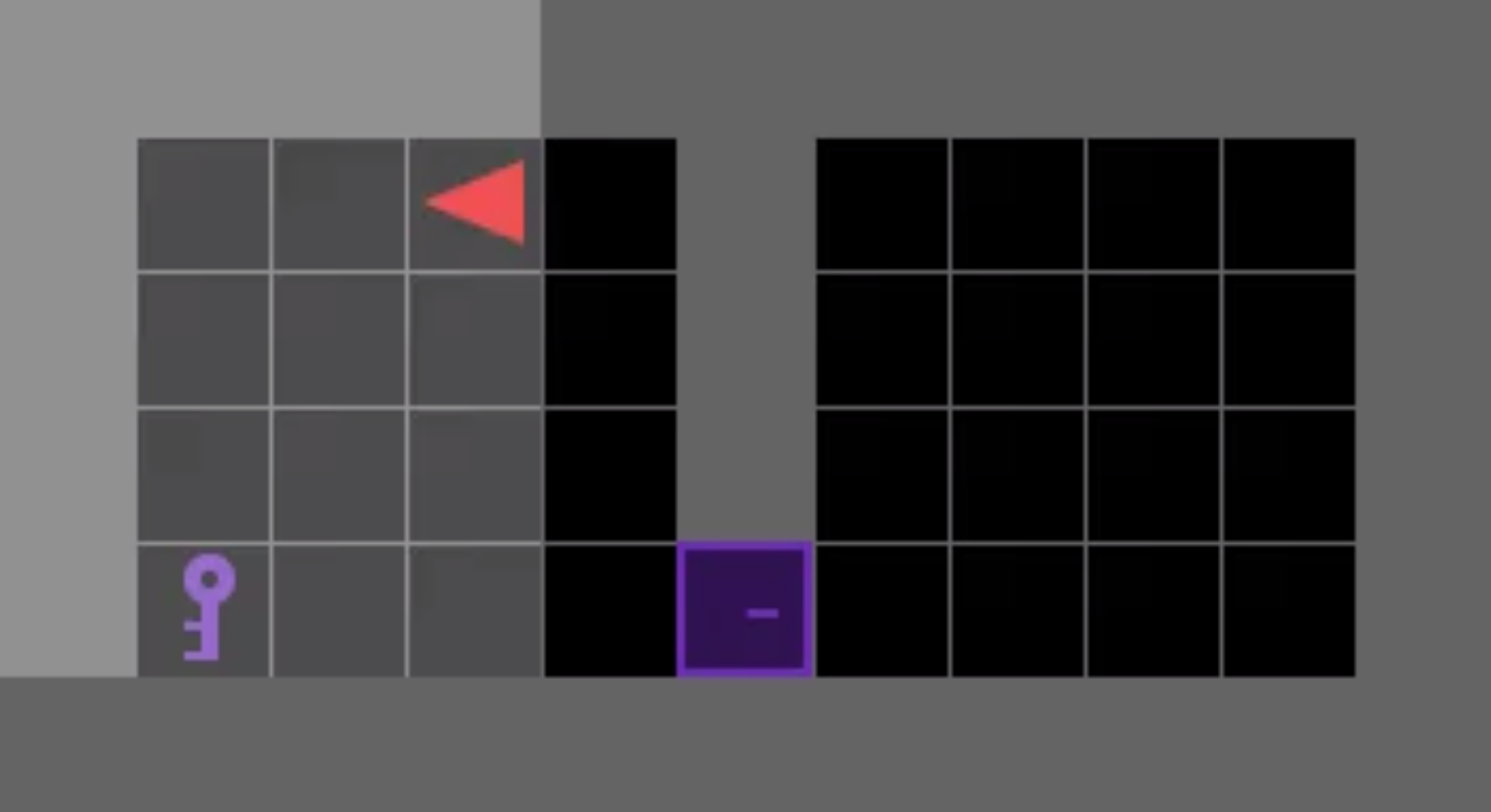} & 
\end{tabular}
\begin{tabular}{l | c c c c | c c c c} 
 \hline
  & \multicolumn{4}{|c}{Average Success Rate} & \multicolumn{4}{|c}{Average Episode Length} \\
         & R6x6 & MR & Unl & Mem & R6x6 & MR & Unl & Mem \\ 
 \hline\hline
 DQN & $99.50$ & $0.00$ & $0.00$ & $40.0$ & $9.31$ & $100.0$ & $100.0$ & $41.14$\\ 
 \hline
 RnD & $100.00$ & $90.00$ & $100.0 $& $48.5$ & $3.50$ & $31.46$ & $4.08$ & $2.78$\\ 
 \hline
 BeBold DQN-CNT & $100.00$ & $98.00$ & $100.0$ & $48.0$ & $3.98$ & $23.51$ & $4.46$ & $2.92$\\
 \hline
 CogNGen & $100.00$ & $98.50$ & $100.0$ & $98.5$ & $3.90$ & $23.41$ & $4.15$ & $2.96$\\ 
 \hline
\end{tabular}
\end{center}
\caption{In the top row, examples of several tasks are presented -- from left to right, the 6$\times$6 empty room task (R$6\times6$), the multi-room task w/ three rooms of size four (MR), and the unlock task (Unl).
In the bottom row, we present results over the last $100$ episodes for: (Left) Average success rate (\%); (Right) Average episode length (\% of maximum episode length - closer to $0$ is more efficient)}
\label{table:task_results}
\vspace{-0.5cm}
\end{table}

\subsection{The Mini GridWorld Problem}
\label{sec:gridworld}

To evaluate CogNGen-built agents, we adapt the environment from the OpenAI Gym extension, Mini-GridWorld \cite{gym_minigrid} and investigate four tasks: the \emph{random empty room}, \textit{multi-room}, \textit{unlocking}, and \emph{memory} tasks. The maze environment is an $N \times M$ tile grid and is partially observable by the agent as a $7 \times 7 \times 3$ tensor created by mapping each tile of the $7\times7$ grid to $3$ integer values. Each tile is encoded to an object index ($0 =$ unseen, $1 =$ empty, $2 =$ wall, etc.), a color index ($0 =$ red, $1 =$ green, etc.), and a state index ($0 =$ open, $1 =$ closed, $2 =$ locked).

The agent itself is restricted to picking up one single object, such as a key, and may open a locked door if it carries a key that matches the door's color. The discrete action space for our agent can be summarized as a set of six unique actions: 1) turn left, 2) turn right, 3) move forward, 4) pick up an object, 5) drop the object that the agent is currently carrying, and 6) toggle/activate (such as opening a door or interacting with an object).
The reward structure/signal provided by all problems in the Mini-GridWorld environment is sparse -- $1.0$ if the agent reaches the green goal tile and $0$ otherwise, making all problems difficult from a reinforcement learning perspective. Each problem has a specific time step limit allotted to allow the agent to complete the task with maximum episode lengths ranging from $60$ to $288$ time steps.

\begin{figure}[!t]
    \centering
    \subfloat[R6$\times$6 task - reward.\label{fig:random_reward}]{%
      \includegraphics[height=1.5in]{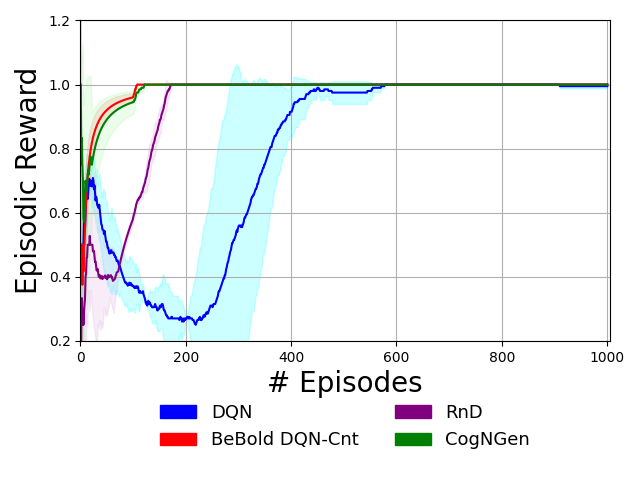}%
    }
     \subfloat[R6$\times$6 - episode length.\label{fig:random_length}]{%
      \includegraphics[height=1.5in]{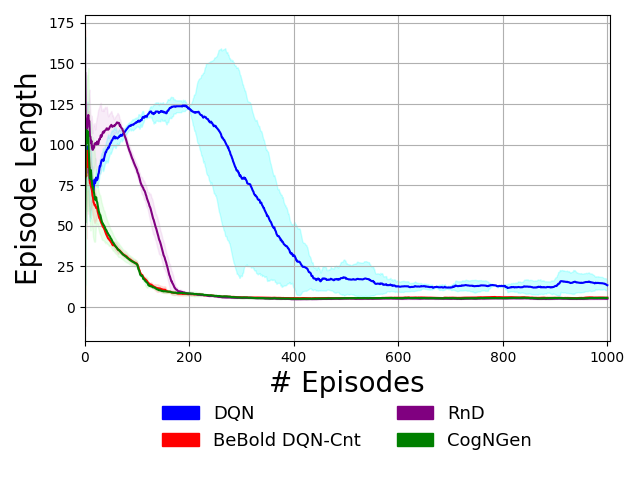}%
    }\\
     \subfloat[MR task - reward.\label{fig:mroom_reward}]{%
      \includegraphics[height=1.5in]{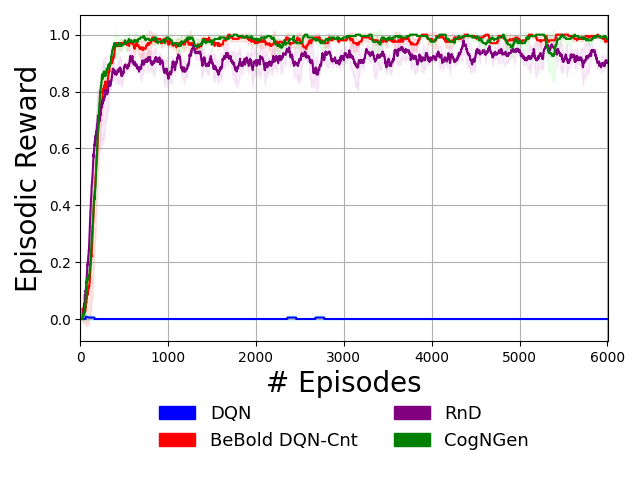}%
    }
     \subfloat[MR task - episode length.\label{fig:mroom_length}]{%
      \includegraphics[height=1.5in]{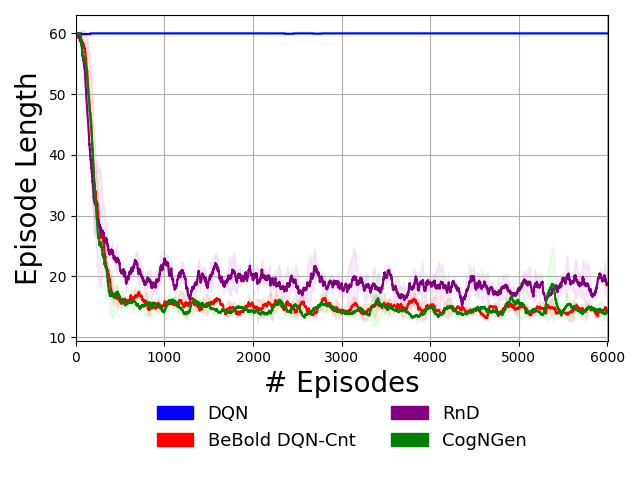}%
    }\\
     \subfloat[Unl task - reward.\label{fig:unlock_reward}]{%
      \includegraphics[height=1.5in]{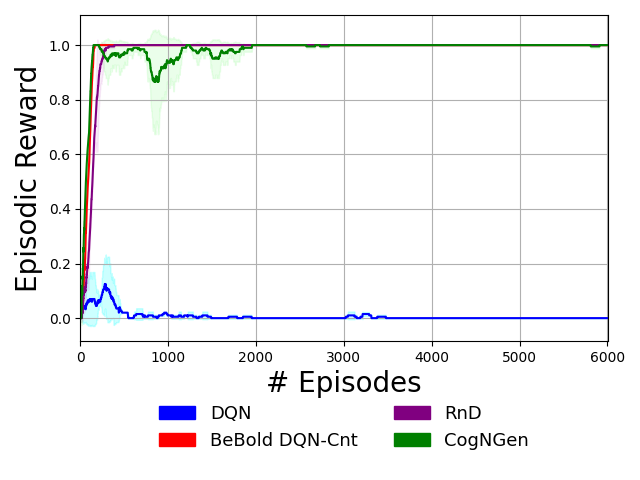}%
    }
     \subfloat[Unl task - episode length.\label{fig:unlock_length}]{%
      \includegraphics[height=1.5in]{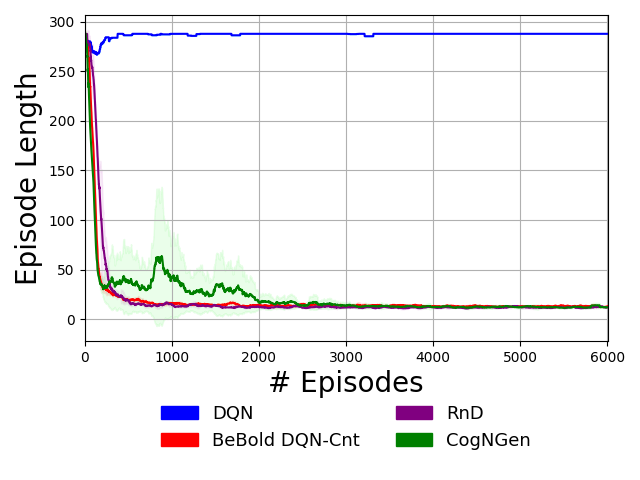}%
    }\\
     \subfloat[Mem task - reward.\label{fig:mem_reward}]{%
      \includegraphics[height=1.5in]{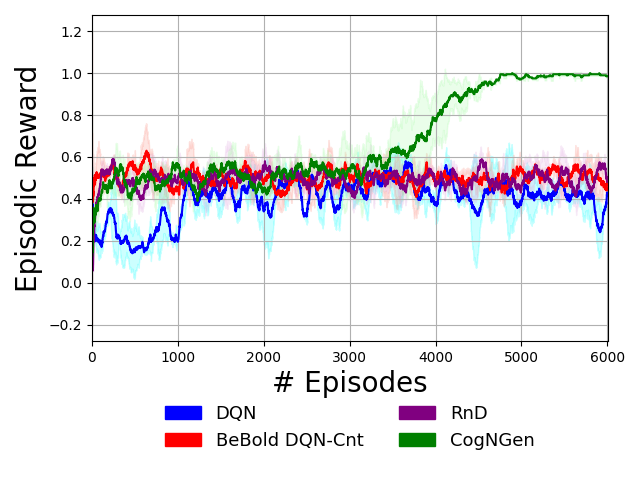}%
    }
     \subfloat[Mem task - episode length.\label{fig:mem_length}]{%
      \includegraphics[height=1.5in]{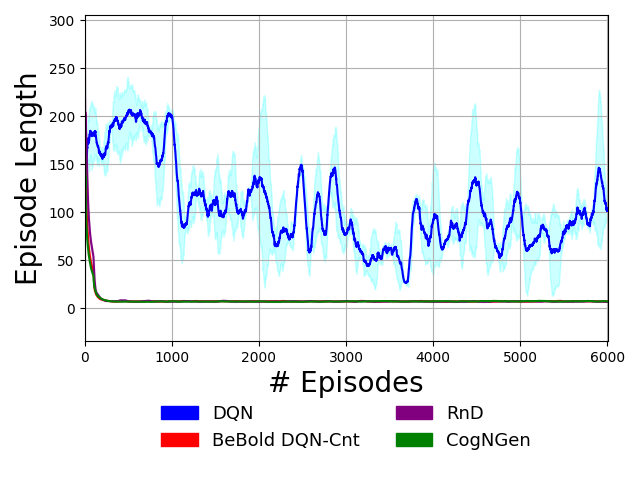}%
    }
    \caption{Average reward (left) and episode length (right) 
    for (top-to-bottom): $6\times6$ empty room (R6$\times$6), multi-room (MR), unlock (Unl), and memory task (Mem).
    }
    \label{fig:task_set1_plots}
    \vspace{-0.55cm}
\end{figure}

\paragraph{The Random Empty Room Task:} In this task (max. $144$ steps), the agent is spawned at a random location (and starting orientation) in the room and must reach the green goal square. A sparse reward is provided if the goal is reached. 

\paragraph{The Multi-Room Task:} This task (max. $60$ steps) requires the agent to navigate a set of connected rooms where it opens a door in one room in order to proceed to the next room. In the final room, there is green square that the agent must reach to end the episode successfully. This is a procedurally generated environment with a different floor plan per episode -- we focused on $3$ rooms of size $4 \times 4$.

\paragraph{The Unlocking Task:}
In this task  (max. $288$ steps), to successfully exit an episode, the agent must open a locked door by finding the key. The key location, door position, and agent initial position/orientation are randomly generated each episode.


\paragraph{The Memory Task:} The agent starts in a small room where it sees an object (such as a key/ball), starting the episode by looking in the direction of the cue object. After perceiving the object, the agent must turn around, exit the room and go through a narrow hall that ends in a split. At the split, the agent can either go up or go down, and at the end of each of these splits is a different object (either a key or ball). To successfully complete the episode (max. $245$ steps) and receive a positive reward, the agent must remember the initial object that it saw and go to the split that contains the correct matching object. 
For this study, we focus on the $7 \times 7$ room variant.

\subsection{Baseline Models}
\label{sec:baselines}

We compare the CogNGen to several baselines: a standard deep Q-network (DQN) \cite{mnih2015human}, a DQN that leverages an intrinsic reward generated via random network distillation (RnD) \cite{burda2018exploration} (an intrinsic curiosity model), and a DQN that learns through a formulation of the BeBold exploration framework \cite{zhang2020bebold} (BeBold DQN-CNT; see Appendix for details). The DQN component of each of the above baselines utilized two layers of hidden neurons 
using the linear rectifier activation. 
RnD and BeBold have access to problem-specific, global information from the Mini GridWorld task environments (namely, the agent's $x-y$ coordinates in the world) whereas CogNGen and the DQN do not. 
For details/hyperparameter settings related to the agent implemented with our CogNGen architecture (referred to as ``CogNGen'' in all plots/tables), please see the Appendix.

\subsection{Experimental Results}
\label{sec:result}

In Table \ref{table:task_results}, we report the average success rate (in solving the task/reaching a goal state) as well as the average episode length (average measurements were computed over the last $100$ episodes of simulation for all models).
In Figure~\ref{fig:task_set1_plots}, we present reward curves (mean \& standard deviation across five trials).

Based on our results, we find that (1) CogNGen is able to learn the maze tasks, (2) the performance is comparable to / on par with powerful deep RL methods that have access to problem-specific, global information, and (3) CogNGen can successfully outperform all baselines on the memory task. Given that CogNGen approximates much of the functionality of modern-day RL mechanisms with large auto-associative Hebbian memory modules and predictive processing circuits, our simulation results are promising. When CogNGen is compared to the baselines, we notice that there are some instances where the powerful BeBold DQN-CNT and RnD baselines yield shorter episodes or yield higher episodic rewards earlier (after converging to an optimal policy). We reason that this small gap is likely due to: 1) BeBold DQN/RnD have access to global, problem-specific information (the agent's $x$-$y$ coordinates in the world in order to calculate state visitation counts) whereas CogNGen only operates with local information, 2) CogNGen's mechanism to update synapses relies on imperfect memory (which is more human-like but introduces error in the recollections as compared to a standard replay buffer), and 3) CogNGen's motor-action model must also learn how to modify its coupled working memory as well as how to interact with its environment, which requires learning more complex policies.




\section{Conclusions}
\label{sec:conclusion}

In this study, we presented CogNGen (the COGnitive Neural GENerative system), a cognitive architecture composed of circuits based on predictive processing and auto-associative Hebbian memory (MINERVA~2). CogNGen lays down the foundation for designing agents composed of neurocognitively-plausible building blocks that learn  across diverse problems as well as potentially model human performance at larger scales. Our results, on a set of sparse reward maze learning tasks, show that goal-directed agents built with CogNGen perform well.
Future work will entail studying the CogNGen's performance on other more complex environments, such as \cite{huang2021gym}, as well as generalizing it further to learning across tasks, i.e., continual reinforcement learning.
\bibliographystyle{splncs04}
\bibliography{ref}

\newpage
\section*{Appendix}
\label{sec:appendix}

In this appendix/supplementary material, we provide details of the predictive processing circuitry that composes several modules of the CogNGen architecture, a general graphical overview/visualization of the CogNGen agent simulated in this study, as well as details of CogNGen's perceptual module, CogNGen's hyper-parameters, and the setup and implementation of the deep network models used as baselines for comparison.

\subsection*{The Neural Generative Coding Circuit}
\label{sec:ngc_circuit}

Neural generative coding (NGC), an instantiation of the predictive processing theory of the brain \cite{rao1999predictive,friston2005theory,clark2015surfing}, is an efficient, robust form of predict-then-correct learning and inference. An NGC circuit in the CogNGen model receives two sensory vectors, input $\mathbf{x}^i \in \mathcal{R}^{I \times 1}$ ($I$ is the input dimensionality) and output $\mathbf{x}^o \in \mathcal{R}^{O \times 1}$ ($O$ is the output or target dimensionality). Compactly, an NGC circuit is composed of $L$ layers of feedforward neuronal units, i.e., layer $\ell$ is represented by the state vector $\mathbf{z}^\ell \in \mathcal{R}^{H_\ell \times 1}$ containing $H_\ell$ total units. Given the input--output pair of sensory vectors $\mathbf{x}^i$ and $\mathbf{x}^o$, the circuit clamps the last layer $\mathbf{z}^L$ to the input, $\mathbf{z}^L =\mathbf{x}^i$, and clamps the first layer $\mathbf{z}^0$ to the output, $\mathbf{z}^0 =\mathbf{x}^o$. Once clamped, the NGC circuit will undergo a settling cycle where it will process the input and output vectors for $K$ steps in times (i.e., it processes sensory signals over a stimulus window of $K$ discrete time steps). 

The activities of the internal neurons themselves (i.e., all of the neurons in between the clamped layers $\ell = L \ldots 0$) are updated according to the following equation (formulated for a single layer $\ell$), in the following manner:
\begin{align}
    \mathbf{z}^\ell \leftarrow  \mathbf{z}^\ell + \beta \Big(-\gamma \mathbf{z}^\ell -\mathbf{e}^\ell + (\mathbf{E}^\ell \cdot \mathbf{e}^{\ell-1}) \otimes \frac{\partial \phi^\ell(\mathbf{z}^\ell)}{\partial \mathbf{z}^\ell}  + \Phi(\mathbf{z}^\ell) \Big) \label{eqn:state_update}
\end{align}
where $\mathbf{E}^\ell$ is a matrix containing error feedback synapses that are meant to pass mismatch signals/messages from layer $\ell-1$ to $\ell$. Although these synapses can be learned, we chose to set it to be $\mathbf{E}^\ell = (\mathbf{W}^\ell)^T$.
$\beta$ is the neural state update coefficient (typically set according to $\beta = \frac{1}{\tau}$, where $\tau$ is the integration time constant in the order of milliseconds) and $\Phi(.)$ is a special lateral interaction function, which we do not use in this work, i.e., $\Phi(\mathbf{v}) = 0$. This update equation indicates that a vector of neural activity changes, at each step within a settling cycle, according to (from left to right), a leak term/variable (the strength of which is controlled by $\gamma$), a combined top-down and bottom-up pressure from mismatch signals in nearby neural regions/layers, and an optional lateral interaction term. $\mathbf{e}^\ell \in \mathcal{R}^{H_\ell \times 1}$ are an additional set/population of special neurons that are tasked entirely with calculating mismatch signals at a layer $\ell$, i.e., $\mathbf{e}^\ell = \mathbf{z}^\ell - \mathbf{\bar{z}}^\ell$, the difference between a layer's current activity (or clamped value) and an expectation/prediction produced from another layer. Specifically, the layer-wise prediction made is $\mathbf{\bar{z}}^\ell$ and is computed as follows: 
$\mathbf{\bar{z}}^\ell = g^\ell( \mathbf{W}^{\ell+1} \cdot \phi^{\ell+1}( \mathbf{z}^{\ell+1} ) )$ where $\mathbf{W}^\ell$ denotes a learnable matrix of generative/predictive synapses. $\phi^{\ell+1}$ is the activation function (which we set to be the linear rectifier in this work) for the state variables and $g^{\ell}$ is a nonlinearity applied to predictive outputs (which we set to be the identity in this study). 

After processing the input--output pair for $K$ time steps (repeatedly applying Equation \ref{eqn:state_update} $K$ times), the synapses are adjusted with a Hebbian-like update:
\begin{align}
    \Delta \mathbf{W} &= \mathbf{e}^\ell \cdot (\phi^{\ell+1}( \mathbf{z}^{\ell+1}) )^T \odot \mathbf{M}_W \label{eqn:predictor_update} \\
    \Delta \mathbf{E} &= \gamma_e (\Delta \mathbf{W})^T \odot \mathbf{M}_E \label{eqn:error_update} 
\end{align}
where $\gamma_e$ is a factor (less than one) to control the time-scale that the error synapses are evolved (to ensure they change a bit more slowly than the generative synapses). $\mathbf{M}_W$ and $\mathbf{M}_E$ are modulation matrices that perform a form a synaptic scaling that ultimately ensures additional stability in the learning process (see \cite{ororbia2021adapting} for details). All NGC circuits in this work are implemented according to the mechanistic process described in this section.

Another important functionality of an NGC circuit is the ability to ancestrally project a vector through the underlying directed generative model. In other words, this amounts to a feedforward pass, since no settling process is required -- we will represent this functionality as $f_{proj}(\mathbf{x}^i; \Theta)$. Formally, ancestrally projecting a vector $\mathbf{x}^i$ through an NGC circuit is done as follows:
\begin{align}
    \mathbf{z}^\ell = \mathbf{\bar{z}}^\ell &= g^\ell( \mathbf{W}^{\ell+1} \cdot \phi^{\ell+1}( \mathbf{z}^{\ell+1} ) ), \; \forall \ell = (L-1),...,0
\end{align}
where $\mathbf{z}^L = \mathbf{x}^i$, i.e., the input (top-most) layer of the circuit is clamped to a specific vector, such as current input pattern $\mathbf{x}^i$).

\begin{figure}[!t]
\begin{center}
\includegraphics[width=0.80\linewidth]{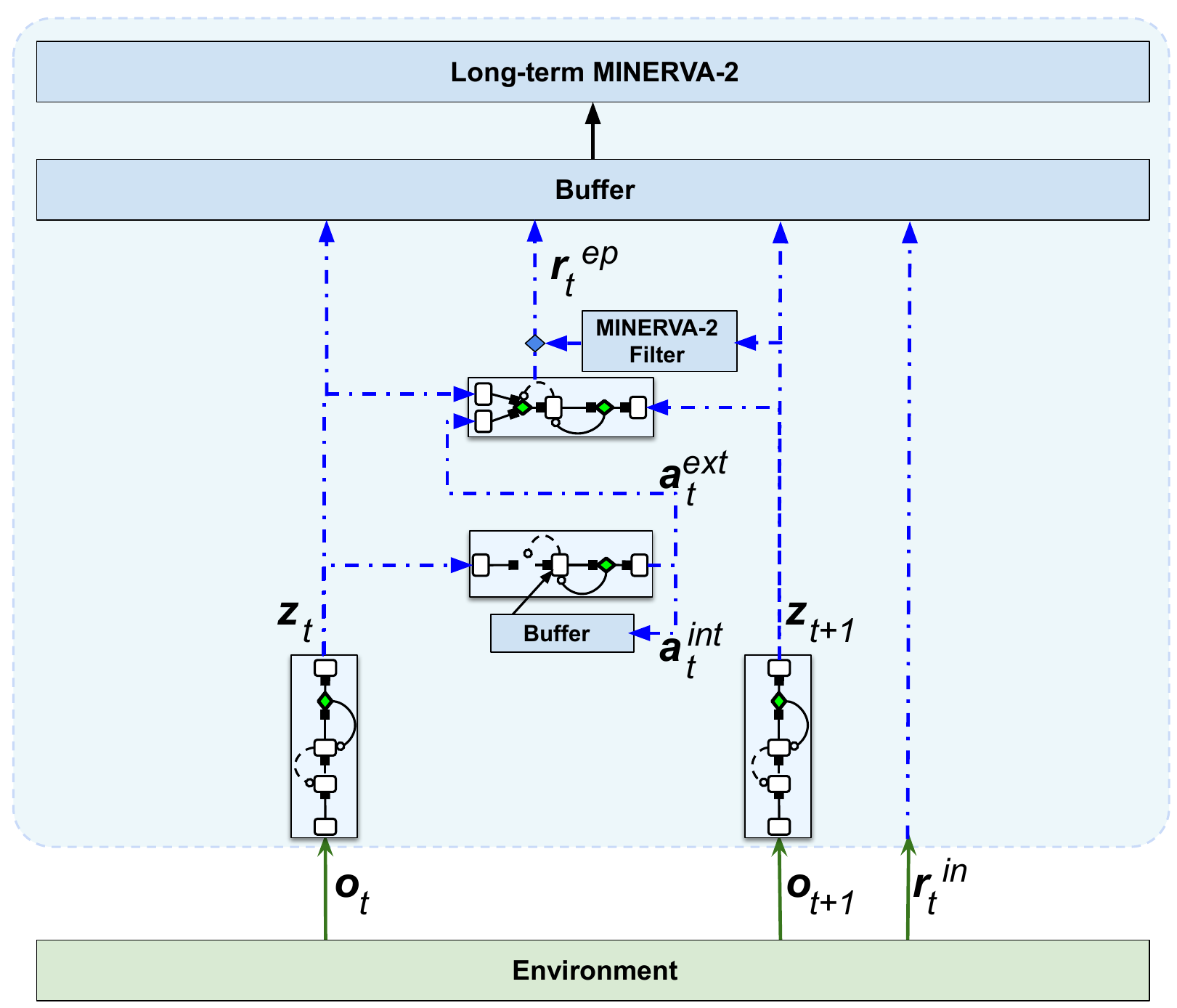}
\end{center}
\caption{The CogNGen maze-learning architecture.
}
\label{fig:cogngen_arch}
\vspace{-0.5cm}
\end{figure}

\subsection*{On the Architecture of CogNGen}

In Figure \ref{fig:cogngen_arch}, we depict the architecture of the CogNGen agent that was simulated for the experiments of this study. Note that this particular diagram depicts CogNGen in ``processing mode'', i.e., no synaptic update are performed in this mode (only circuit latent states are updated, actions are taken, and internal reward signals are computed/filtered). However, relevant information is stored in a buffer (serving as working memory) which interfaces with the long-term MINERVA-2 memory. Transitions are sampled from the MINERVA-2 memory module when the ``learning model'' is triggered, i.e., in this work, after $\mathbf{z}_{t+1}$ has been encountered by the perceptual module and stored, samples are replayed from memory to update the motor and procedural memory (dynamics) circuits. 
Note that while Figure \ref{fig:cogngen_arch} depicts the general case where the perceptual module is an NGC circuit itself, in this work, we replaced this learnable circuit with the Gym-MiniGrid problem specific encoder/decoder (to speed up simulation).

\paragraph{On the Perceptual Modules:} In general, an NGC circuit can be constructed to serve the role of the encoder $f_e$ for CogNGen. Doing so would yield the additional advantage that a top-down directed generative model, or decoder $f_g \colon \mathbf{z}_t \mapsto \mathbf{o}_t$, would be learned implicitly with $f_e$, given that, in prior work, we have shown that NGC learns a good density estimator of data (from which new samples can be ``fantasized'' or synthesized) \cite{ororbia2020continual,ororbia2022neural}. An NGC encoder would, by default, be unsupervised, especially if it is being pre-trained before a task's simulation, i.e., the NGC $f_e$ would be trained to predict $\mathbf{o}_t$ given $\mathbf{z}_t$, where $\mathbf{z}_t$ is iteratively crafted by the NGC settling process described in Section \ref{sec:ngc} (and a feedforward model can be trained to approximate and amortize the settling process to further reduce computational complexity). Having a decoder would also allow for visual interpretation of the distributed representations acquired by CogNGen since a latent vector $\mathbf{\hat{z}}_t$ (such one produced by the procedural dynamics model, described later) could be run through the underlying top-down directed generative $f_g$ to produce its corresponding instantiation $\mathbf{\hat{o}}_t$ in observation space.

Another advantage of the encoder formulation is that if a  task-specific (pre-trained/pre-designed) encoder $f_e$ for a given modality is available, it may be utilized alongside or in place of the NGC encoder circuit describe above. This can simplify and speed up the simulations involving CogNGen, especially if learning a joint perceptual-memory-control system is not the goal, allowing the experimenter to leverage a reliable, stable state representation to design or experiment with various configurations/alterations of the CogNGen kernel's other internal sub-systems and observe their impact on the task at hand. Note that in this paper, we opted to utilize the world-specific encoders/decoders that were provided with the Gym-MiniGrid environment and will explore learning the NGC circuit encoders/decoders in future work.

\paragraph{CogNGen Agent Details: Hyperparameters:} The specific instantiation of CogNGen we simulated for the experiments conducted for this study utilized the following hyper-parameter settings: Both motor and procedural memory/dynamics circuits were optimized with the Adam update rule with a learning rate of $0.0005$. The motor model contained two hidden/latent layers of $512$ neurons while the procedural circuit contained two layers of $128$ neurons -- both circuits used the linear rectifier, i.e., $max(0,v)$, as the activation function. For the motor model, a discount factor of $\gamma = 0.99$ was used and its underlying target network (which was used in order to improve stability of its bootstrap estimation process) was updated to match the current motor model's synaptic weight values every $128$ transitions. Although the motor model makes use of an epistemic signal to drive useful/intelligent exploration, a light epsilon-greedy action scheme was used at the start of the learning process where a random action was selected with probability $\epsilon$ -- this was rapidly decayed from $\epsilon = 0.95$ to $\epsilon = 0.0$ for external actions (for all tasks) and held at $\epsilon = 0$ for internal actions except for the multi-room (MR) task, where it was rapidly decayed $\epsilon = 0.1$ to $\epsilon = 0.0$.

The working self-recurrent slot buffers were set to contain two slots each with an embedding dimension of $100$ recurrent neurons. Both the long-term and short-term/working MINERVA~2 memory models were set to use a power of $100$ and the long-term memory processed sequence chunks with window length of $10$. The total size of the long-term MINERVA~2 was bounded at a maximum of $10^6$ memories and mini-batches of $256$  transitions were sampled from it whenever the motor and procedural circuits were to be updated.

\subsection*{Baseline Implementation Details}

We compare the CogNGen to several baselines: a standard deep Q-network (DQN) \cite{mnih2015human}, a DQN that leverages an intrinsic reward generated via random network distillation (RnD) \cite{burda2018exploration} (an intrinsic curiosity model), and a DQN that learns through a formulation of the BeBold exploration framework \cite{zhang2020bebold} (BeBold DQN-CNT). The DQN component of each of the baseline models utilized two layers of hidden neurons  (the size of each were searched in the range of $[128, 512]$) 
using the linear rectifier activation. 
For RnD, the predictor $\hat{f}(\mathbf{z}_{t+1})$ and random target network $f(\mathbf{z}_{t+1})$ both contained two layers of neurons (size of which was searched in the range of $128$ through $512$) also using the linear rectifier activation. The weight parameters for all DQNs as well as the RnD's predictor and random networks were initialized  according to the scheme in \cite{he2015delving} and parameters were optimized by calculating gradients using reverse-mode differentiation and the Adam adaptive learning rate with step size searched in the range $0.0002$ through $0.001$.
The BeBold DQN-CNT utilized global state visitation counts to compute its intrinsic reward bonus (meaning that we implemented and tuned the ``episodic restriction on intrinsic reward'', or ERIR, model in \cite{zhang2020bebold}). We calculate the intrinsic reward for the BeBold model as follows:
\begin{align*}
    r^i &= \max\Big( 0, \frac{1}{\mathds{N}(\mathbf{z}_{t+1})} - \frac{1}{\mathds{N}(\mathbf{z}_{t})} \Big) \Big( \mathbbm{1}\Big\{ \frac{1}{\mathds{N}_e(\mathbf{z}_{t+1})} \Big\} \Big) \\
    r^i_t &= ( r^i > 0 \rightarrow r^i ) \land ( r^i \leq 0 \rightarrow -\alpha )
\end{align*}
where $0.1 \leq \alpha \leq 1$ and $\mathds{N}(\mathbf{z}_t)$ is the hash table that returns the global visitation of state $\mathbf{z}_t$ while $\mathds{N}_e(\mathbf{z}_t)$ returns the episodic visitation count of $\mathbf{z}_t$. Note that the key needed to retrieve is the count value is the $x$-$y$ coordinate of state $\mathbf{z}_t$ extracted from the problem environment. For RnD, our implementation of the intrinsic reward proceeded as follows:
\begin{align*}
    r^i &= \Big( ||\hat{f}(\mathbf{z}_{t+1}) - f(\mathbf{z}_{t+1})||_2^2 \Big) \Big( \mathbbm{1}\Big\{ \frac{1}{\mathds{N}_e(\mathbf{z}_{t+1})} \Big\} \Big) \\
     r^i_t &= ( r^i > 0 \rightarrow r^i ) \land ( r^i \leq 0 \rightarrow -\alpha )
\end{align*}
In order to obtain robust and stable performance, we had to modify the RnD and BeBold intrinsic bonus calculations in order to learn in the above tasks by imposing a small negative penalty on discrete states that were visited more than once within an episode (meaning that a hash table had to be used to track the global state coordinates and visitation counts of each prior state seen, which was reset at the end of each episode). 
As noted in the main paper, the RnD and BeBold baselines had access to problem-specific, global information from the Mini GridWorld task environments (namely, the agent's $x-y$ coordinates in the world) whereas CogNGen and the DQN do not. This was found to be necessary to obtain good performance from these baselines (if the global count information was removed, both the RnD and BeBold models struggled to perform consistently well).

\end{document}